\documentclass[10pt,twocolumn,letterpaper]{stanford}

\usepackage{bbm}
\usepackage[pagenumbers]{cvpr} % To force page numbers, e.g. for an arXiv version
\usepackage{tcolorbox}
\usepackage{longtable}
\definecolor{cvprblue}{rgb}{0.21,0.49,0.74}
\definecolor{stanfordblue}{HTML}{006CB8}
\definecolor{todo}{rgb}{0.9,0.1,0.1}
\definecolor{row_color}{HTML}{F3F8FC}
\definecolor{human_color}{rgb}{0.9,0.9,0.9}
\usepackage[pagebackref,breaklinks,colorlinks,allcolors=stanfordblue]{hyperref}

\usepackage{xcolor}
\usepackage{booktabs}
\usepackage{colortbl}
\usepackage{multirow}
\usepackage{caption}
\usepackage{balance}
\usepackage{flushend}

\definecolor{QuantiDark}{RGB}{169,183,163}   % example: dark green for best overall
\definecolor{QuantiLight}{RGB}{227,232,224} % example: light green for best open
\definecolor{QuantiDarkYellow}{RGB}{227,209,169}   % example: dark yellow for best overall
\definecolor{QuantiLightYellow}{RGB}{249,242,228} % example: light yellow for best open
\definecolor{QuantiDarkRed}{RGB}{181,106,69}   % example: dark red for best overall
\definecolor{QuantiLightRed}{RGB}{239,225,216} % example: light red for best open
\definecolor{QuantiDarkBlue}{RGB}{34,59,61}   % example: dark red for best overall
\definecolor{QuantiLighBlue}{RGB}{210,216,215} % example: light red for best open

\def\paperID{****} % *** Enter the Paper ID here
\def\confName{CVPR}
\def\confYear{2026}

\addtocontents{toc}{\protect\setcounter{tocdepth}{-1}}

\usepackage[draft]{minted}

\usepackage{xspace}
\newcommand{\ours}{\textsc{OpenWAM}\xspace}

\title{
    \textsc{OpenWAM}: An Open Framework for Composable World-Action Models
}
\author[*]{\text{Heng Yu}}
\author[*]{\text{David D. Yuan}}
\author[*]{\text{Juze Zhang}}
\author{\text{Changan Chen}}
\author{\text{Yao Feng}}
\authorbreak
\author{\text{Michelle Baldonado}}
\author{\text{Steve Cousins}}
\author{\text{Li Fei-Fei}}
\author{\text{Jiajun Wu}}
\author{\text{Ehsan Adeli}}
\affiliation{Stanford University}
\contribution{\textsuperscript{*}Equal contribution}
\metadata[Project Page]{\url{https://openwam.stanford.edu}}
\metadata[Code]{\url{https://github.com/OpenWAM/OpenWAM}}
\metadata[Release]{Project page \& code: June 4, 2026 \qquad Paper: October 6, 2026}

\begin{document}

\abstract{

World--action models (WAMs) couple future prediction with robot control, yet
existing systems often vary the video backbone, interaction structure,
supervision, and inference procedure simultaneously, making their design
choices difficult to compare. We introduce \textbf{\ours}, an open
world--action modeling framework built around a common causal robot-video
foundation and configurable video--action interaction.
Starting from Wan2.2-5B, we perform causal robot-video pretraining on over 10,000 hours of video, then integrate an action expert through a shared Mixture-of-Transformers architecture that supports joint, video-then-action, action-then-video, and decoupled generation.
\ours achieves high success rates on four LIBERO suites and real-world bimanual tasks; robot-video training with causal adaptation improves VTA success on LIBERO-Long from 68.4\% to 97.8\%.
The same configurable architecture naturally extends to inverse and forward
dynamics, allowing us to study how counterfactual transitions improve
independently trained dynamics models beyond demonstrations alone.
When only the video predictor is adapted to a new task, a frozen local-context
inverse dynamics model trained on counterfactual data and demonstrations achieves 84.0\% mean success across four held-out
LIBERO-90 tasks, compared with 47.0\% for a full-context inverse model and
21.5\% for a local-context model trained only on demonstrations.
For forward dynamics, counterfactual supervision reduces RGB prediction error by
34.5\% and raises outcome identification from 21.1\% to 71.3\% among 16
same-state outcomes.
\ours provides a common testbed for comparing WAM interaction designs and for studying dynamics learning from video data beyond successful demonstrations.
}

\maketitle

\section{Introduction}

World-action models (WAMs) integrate future-observation prediction with robot control. Recent systems such as DreamZero~\cite{ye2026world}, DVA~\cite{rhoda2026dva} and LingBot-VA~\cite{li2026causal, lingbotva2} use pretrained video models to couple future-observation prediction with robot action generation. However, differences in their backbones, training data, objectives, and inference procedures make it difficult to isolate the effect of video--action interaction.

\begin{figure*}[t]
  \centering
  \includegraphics[width=0.98\textwidth]{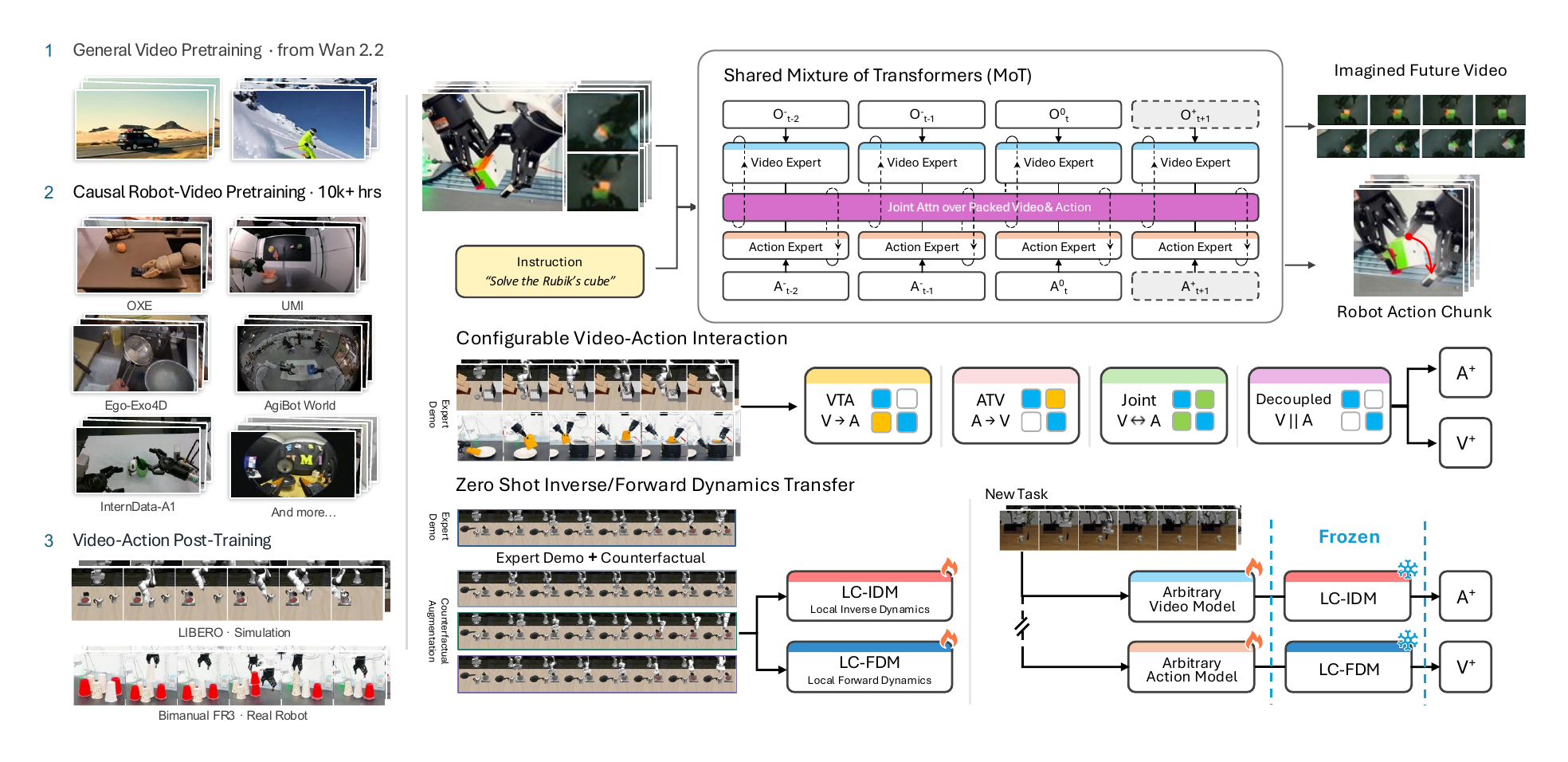}
  % \vspace{-0.3in}
  \caption{
\textbf{Overview of \ours.}
\textbf{Left:} Starting from Wan2.2~\cite{wan2025wan}, causal pretraining on robot and
human-interaction video builds a predictive backbone without action labels,
followed by video--action training on simulated and real-robot demonstrations.
\textbf{Top right:} A shared Mixture-of-Transformers couples temporally
aligned video and action streams through joint attention while retaining
modality-specific experts.
\textbf{Middle right:} Configurable cross-modal attention and generation
order support joint, video-then-action, action-then-video, and decoupled
prediction within the same architecture.
\textbf{Bottom right:} Local-context inverse and forward dynamics models
are trained independently using demonstrations and counterfactual
transitions. A frozen IDM translates predictions from compatible
task-adapted video models into actions, while an FDM predicts the visual
consequences of candidate actions.
}
  \label{fig:openwam-teaser}
\end{figure*}

Two challenges arise when studying these systems systematically. First, video--action interaction spans multiple choices: video and actions can be generated jointly, sequentially, or through restricted cross-modal connections. A shared architecture and evaluation protocol would allow these interaction patterns and generation orders to be compared under controlled conditions.
Second, successful demonstrations cover only a limited subset of possible action--outcome relationships. Other interactions, including failures, exploratory behavior, and counterfactual rollouts, can provide additional supervision about local dynamics. However, it remains unclear whether such data can improve the reuse of an action-grounding component when the visual predictor is adapted to a new task. We study this question by independently training local-context dynamics models and evaluating their transfer across adapted video predictors.

To this end, we introduce \textbf{\ours}, an open framework and model series for controlled world--action modeling, summarized in Figure~\ref{fig:openwam-teaser}. Starting from a pretrained general-purpose video model, we adapt its visual prior through causal robot-video training without action labels, then introduce an action branch through a Mixture-of-Transformers (MoT) architecture~\cite{liang2024mixture}. The resulting model supports joint, sequential, and restricted video--action interaction. These configurations share the video initialization, representations, and evaluation pipeline, allowing us to compare how their interaction patterns affect the control performance.

Furthermore, \ours trains \textbf{local-context inverse and forward models} independently. The inverse model predicts actions from the current observation and a future visual trajectory, while the forward model predicts that trajectory from supplied actions.
In simulation, we broaden their training data by executing alternative action sequences from the same initial state.
% On physical robots, we supplement inverse-model training with failed attempts and free-form play.
This local-transition formulation is also compatible with failed and exploratory interactions, because it does not require a successful task outcome or a success label.
Each recorded transition supplies dynamics supervision regardless of whether it completes the task. At inference, a video predictor can propose a visual trajectory and a separately trained inverse model can translate it into executable actions.

We evaluate robot-video pretraining, closed-loop policy success across
video--action interaction modes and dynamics-component reuse. In transfer experiments, task-adapted video predictors are paired with frozen inverse models to compare demonstration-only with counterfactual-augmented supervision and local-context with full-context conditioning. We additionally evaluate forward prediction under alternative actions from the same initial state.

This work makes three contributions:
\begin{itemize}
  \item \textbf{A controlled WAM framework.} \ours combines a causal video backbone with configurable video--action models under shared training and evaluation protocols.
  \item \textbf{Local-context dynamics interfaces.} We independently train inverse- and forward-dynamics models from demonstrations and broader local transitions and evaluate their composition with separately trained predictors.
  \item \textbf{Controlled empirical analysis.} We measure the effects of robot-video pretraining, interaction design, transition coverage, and conditioning context on policy performance and component reuse.
\end{itemize}

\section{Related Work}
\label{sec:related}

\subsection{Visuomotor Policies and Predictive Control}

Behavior cloning provides the standard supervised formulation for visuomotor
policy learning, mapping observations and task context directly to demonstrated
actions. Action chunking and generative action prediction have become effective
control primitives~\cite{zhao2023learning, chi2025diffusion}, while large-scale
vision--language--action models extend this paradigm through language
conditioning, pretraining, and cross-task or cross-embodiment learning
\cite{brohan2022rt, brohan2023rt, kim2024openvla, black2024pi_0,
intelligence2025pi_, o2024open, team2024octo}.

Predictive control introduces an additional interface by explicitly modeling
future observations or latent states. Early visual-foresight methods combined
action-conditioned video prediction with model-predictive control for robotic
manipulation~\cite{finn2017deep,ebert2018visual}. Subsequent work learns latent
or visual dynamics to support policy learning, planning, and imagined rollouts
\cite{hafner2019dream, hafner2023mastering, wu2023daydreamer,
wu2024unleashing, cheang2024gr, zhou2024robodreamer}. Predictive rollouts have also been used to evaluate policy behavior~\cite{li2025worldeval, guo2026ctrl}.

Action-conditioned world models provide a complementary perspective by
predicting how an environment evolves under supplied or candidate actions.
Recent methods use such predictions for simulation, candidate evaluation,
proposal refinement, and self-verifying planning
\cite{huang2026learning, zhou2026tau_0, qi2026inference,
zhang2026worldcoherent}. Together, these lines show that predictive models can
support action generation, simulation, planning, and evaluation. However,
action-conditioned prediction and video--action policy generation are often
developed as separate components, leaving open how explicit
prediction--action interfaces affect component reuse across tasks.

\subsection{Video Foundations and World--Action Architectures}

Recent world--action models connect predictive video representations with
robot control through several forms of video--action coupling. Joint
generation models predict future visual content and actions together
\cite{ye2026world, cosmos3, shen2026videovla, guo2024prediction}, while
causal video-first formulations generate or represent visual futures before
action prediction~\cite{li2026causal, rhoda2026dva}. Other architectures use
specialized video and action pathways with explicit cross-modal exchange
\cite{yang2026covar, zhang2026deva}, or leverage egocentric and large-scale
video priors for robot grounding and generalization
\cite{li2026egowam, zhou2026zero, wu2026zimablue, qiu2026vid2wam}.

A closely related line studies the interface between predictive visual
information and executable actions. Predicted visual states or representations
can condition inverse dynamics~\cite{tian2025predictive, hu2024video, li2026turningvideomodelsgeneralist}, while
other approaches translate generated futures through dense correspondence,
pseudo-action labeling, intermediate denoising features, or latent
representation alignment
\cite{ko2024learning, jang2025dreamgen, ma2026dit4dit,
vuong2026world2act}. These methods differ in where and how predictive visual
information enters the action-generation pathway.

Unified formulations instead expose multiple predictive and control modes
within a shared system. UWM and UVA~\cite{zhu2025unified, li2503unified}
support combinations of policy, inverse-dynamics, forward-dynamics, and video
prediction objectives, while related systems combine action generation with
world prediction, simulation, planning, or evaluation
\cite{cen2025worldvla, kim2026cosmos, sun2026riemann, bi2026motus2,
guo2026unified}.

Recent work further isolates particular choices in the WAM design space.
Predictive computation can be reduced, sparsified, reordered, or scheduled
asynchronously
\cite{yuan2026fast, tang2026world, wang2026glancewam,
ye2603gigaworld, li2026light, huang2026streaming,
huang2026noisegate, pan2026selfwam}. Alternative work studies the
representation through which world prediction interacts with control
\cite{liu2026oa, wang2026repwam, qiu2026making, zhao2026sg,
shen2026ld4wam, you2026affordancewam, jiang2026rethinking}, or augments
world--action learning with structural, geometric, and tactile supervision
\cite{zheng2026gift, gonzalez2026spatially, ma2026tacpac}. Temporal
abstraction, data and model scaling, humanoid control, and long-horizon memory
provide additional design axes
\cite{li2026wall, liu2026ge, zheng2026motionwam, zhao2026memory};
recent surveys summarize the rapidly expanding landscape
\cite{wang2026world, lu2026world}.

Concurrent work, also named OpenWAM~\cite{wang2026openwam}, develops a modular
framework for systematic WAM pretraining and design studies. Our focus is
complementary: \ours fixes a common causal robot-video foundation and training
interface while explicitly varying generation order and future cross-modal
visibility, enabling joint, video-then-action, action-then-video, and decoupled
interaction to be compared under a shared evaluation protocol. We further study
the reuse of independently trained inverse- and forward-dynamics components
through explicit trajectory interfaces.

\subsection{Dynamics Learning Beyond Successful Demonstrations}

Successful demonstrations provide valid transitions, but they cover only the
action--outcome relationships selected by the demonstrating policy. A
complementary analysis shows that observational demonstrations alone may not
identify the consequences of unsupported actions~\cite{yu2026capability}.
Classical imitation-learning methods address the resulting distribution shift
by collecting supervision on learner-visited states, perturbing expert
trajectories, or soliciting corrective interventions
\cite{ross2011reduction,laskey2017dart,mandlekar2020human,hu2026rac,
intelligence2025pi}. These approaches broaden state coverage beyond clean
expert rollouts and improve robustness to execution errors.

More recent work expands supervision through synthetic failures,
counterfactual augmentation, autonomous experience, and other non-successful
interactions
\cite{ameperosa2025rocoda,duan2024aha,li2026learning,zhao2026flare,
yin2026playworld,peng2026fact,bi2026motus2,yang2026wam}.
Such data have been used for recovery, policy improvement, failure
understanding, and predictive modeling. Particularly related to our setting,
Fail2Progress~\cite{huang2025fail2progress} targets data collection toward
observed failures to improve skill-effect models, while
UniPi~\cite{du2023learning} decouples video planning from inverse dynamics and
allows the inverse model to learn from a separate, potentially suboptimal
dataset.

\ours focuses on a different question: how transition coverage and conditioning
scope affect the reuse of independently trained dynamics components after
visual task adaptation. We branch alternative action continuations from
task-relevant simulator states and use the resulting state--action--outcome
transitions as local dynamics supervision.

\section{Methodology}
\label{sec:method}

We develop \ours as an open framework for composing world prediction and
action generation. It combines three ingredients: a causal video backbone
pretrained on robot video without action labels, a common video--action
architecture whose cross-modal interaction can be reconfigured, and
separately trained inverse and forward dynamics components operating through
local trajectory interfaces.

The design supports two forms of composition. At the \emph{architectural}
level, we vary how future observation and action streams interact while
retaining a common backbone and objective family. At the
\emph{component} level, we connect independently trained predictors through
explicit future-observation and action-trajectory interfaces. The latter
motivates local-context dynamics models that do not require the language and
nonlocal history used by a task-conditioned generator.

% \subsection{Composable World--Action Modeling}
\subsection{World--Action Modeling Formulation}
\label{sec:world-action-interfaces}

Let \(o_t \in \mathcal{O}\), \(a_t \in \mathcal{A}\), \(q_t \in \mathcal{Q}\), and \(\ell\in\mathcal{L}\) denote the visual observation, action, proprioceptive state, and language instruction at time $t$. For fixed observation- and action-history lengths \(K_o\in\mathbb{N}\) and \(K_a\in\mathbb{N}\), respectively, we define
\begin{equation}
\begin{aligned}
O_t^- &= (o_{t-K_o}, \ldots, o_{t-1}), \\
A_t^- &= (a_{t-K_a}, \ldots, a_{t-1}), \\
O_t^0 &= o_t.
\end{aligned}
\end{equation}

Let \(H\in\mathbb N_{>0}\) denote the prediction horizon. The future action and observation trajectories are defined as
\begin{equation}
% \begin{aligned}
A_t^+ = (a_t, \ldots, a_{t+H-1}), \quad O_t^+ = (o_{t+1}, \ldots, o_{t+H}).
% \end{aligned}
\end{equation}

We collect the information available before predicting the future into the task context
\[
\mathcal C_t=(O_t^-,O_t^0,A_t^-,q_t,\ell).
\]

Recorded future trajectories are denoted without hats, whereas generated trajectories are denoted by \(\hat O_t^+\) and \(\hat A_t^+\). We omit the time index \(t\) when it is unambiguous.

\ours models future observations and actions jointly. For a reference
joint distribution, the two chain-rule factorizations are
\begin{equation}
\begin{aligned}
p_{\mathrm{WAM}}
( O_t^+, A_t^+\mid\mathcal C_t)
&=
\underbrace{p( O_t^+\mid\mathcal C_t)}_{\mathrm{CVP}}
\underbrace{
p( A_t^+\mid O_t^+,\mathcal C_t)
}_{\mathrm{IDM}}\\
&=
\underbrace{p( A_t^+\mid\mathcal C_t)}_{\mathrm{VLA}}
\underbrace{
p( O_t^+\mid A_t^+,\mathcal C_t)
}_{\mathrm{FDM}}.
\end{aligned}
\label{eq:wam-factorizations}
\end{equation}
Causal video prediction (CVP) proposes future observations, a direct
vision--language--action (VLA) policy proposes actions, inverse dynamics
(IDM) grounds a supplied visual transition into actions, and forward
dynamics (FDM) predicts the visual consequence of supplied actions.

These roles give three primary generation routes.
\emph{Joint generation} produces future observations and actions
concurrently. \emph{Video-then-action (VTA)} first generates a visual
future and then conditions action generation on that trajectory.
\emph{Action-then-video (ATV)} first proposes actions and then predicts
their visual consequences. These factorizations specify modeling
interfaces; they do not require all factors to share one trained
parameterization or final checkpoint.

\subsection{An Open Causal Robot-Video Backbone}
\label{sec:causal-video-backbone}

% We initialize the visual backbone from Wan2.2-5B~\cite{wan2025wan} and
% further pretrain it on heterogeneous robot and interaction video without action labels. This stage
% adapts a general video prior toward robot-centric motion and interaction
% while retaining a standalone predictive model. The resulting checkpoint
% provides the common visual initialization for the downstream world--action
% configurations. Dataset composition and optimization settings are given in
% Sec.~\ref{sec:experimental-setup}. We release the robot-video-pretrained
% checkpoint and its training recipe for standalone prediction and downstream
% world--action modeling.

We initialize the video backbone from Wan2.2-5B~\cite{wan2025wan} and
continue pretraining it on heterogeneous robot-manipulation and interaction
videos without action supervision. This stage adapts a general-purpose video
prior to robot-centric motion and interaction while preserving a standalone
video-prediction interface. The resulting checkpoint is shared across all
downstream \ours variants, allowing different video--action interaction
strategies to be compared under a common visual initialization. Dataset
composition and optimization details are summarized in
Sec.~\ref{sec:experimental-setup}.

Because the original video model is not temporally causal, we introduce
chunk-causal prediction during robot-video pretraining. A future chunk may
attend to the current and preceding observations and previously generated
chunks, but not to later chunks; tokens within a chunk are processed
jointly. This provides the temporal direction needed for sequential
prediction while retaining within-chunk denoising.

Temporal causality and cross-modal interaction are separate design
dimensions. Chunk causality determines which times are visible, while the
interaction programs in Sec.~\ref{sec:interaction-programs} determine which
observation and action streams communicate. The pretrained causal backbone
can therefore be used directly as a video predictor or coupled to an action
branch without changing its basic predictive interface.

\subsection{Configurable Video--Action Interaction Programs}
\label{sec:interaction-programs}
\label{sec:framework-architecture}

\begin{figure}[t]
  \centering
  \includegraphics[width=\linewidth]{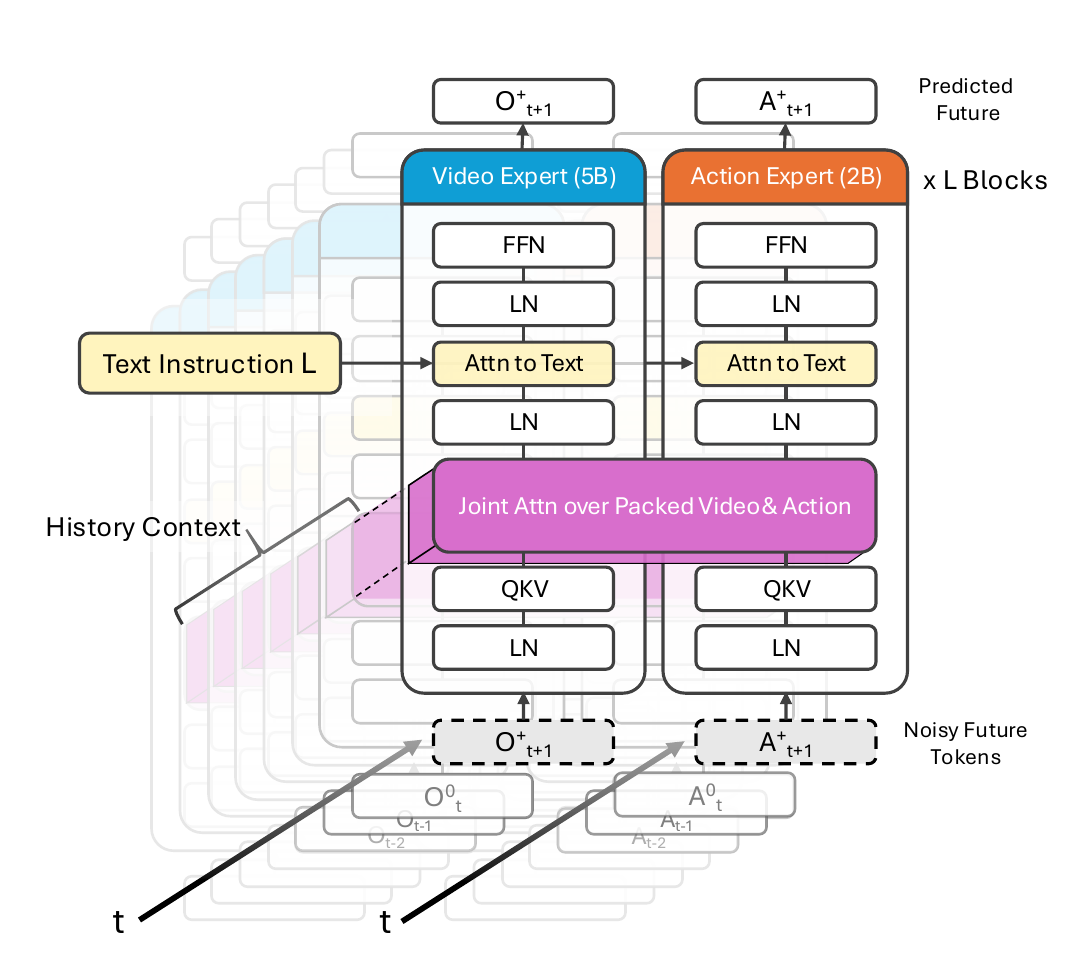}
  % \vspace{-0.4in}
  \caption{
  Shared MoT architecture. Two modality-specific branches keep their own
  normalization, projection and feed-forward parameters and meet in a single
  attention over the packed video and action tokens; a separate
  cross-attention consumes the task text. Noisy future tokens enter at the
  bottom and the preceding chunks sit behind as history context.
  }
  \label{fig:openwam-mot-programs}
\end{figure}

\begin{figure}[t]
  \centering
  \includegraphics[width=0.98\linewidth]{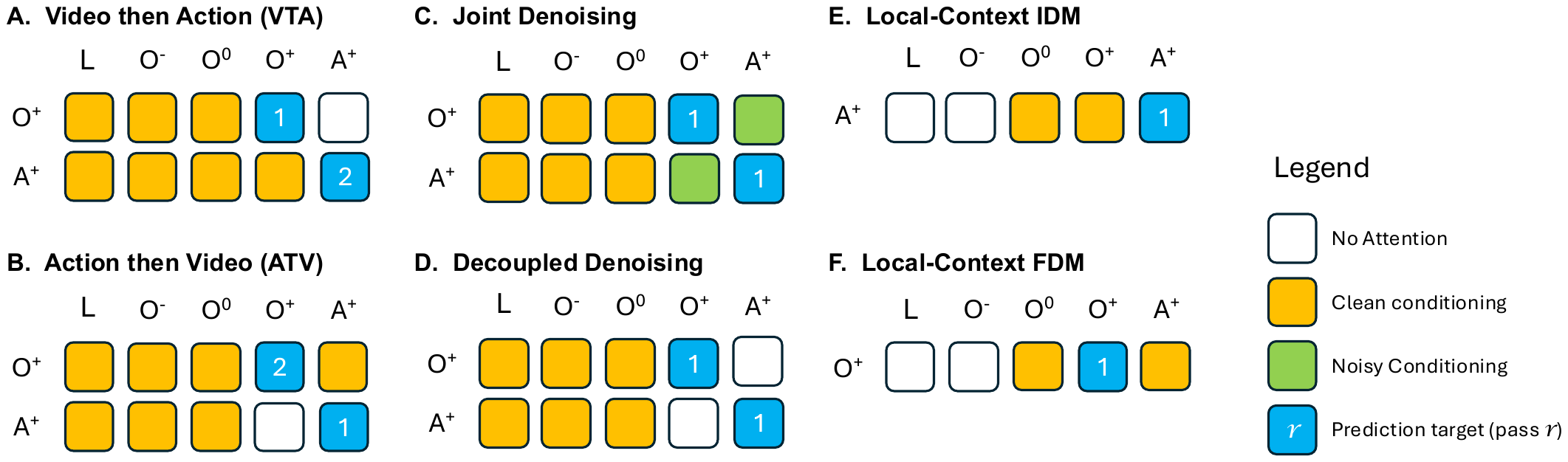}
    % \vspace{-0.1in}
  \caption{
  Attention masks the shared architecture exposes: four interaction programs
  (A--D) and the two local-context dynamics programs (E--F). Columns are the
  streams a row may condition on, cell colour gives what it attends to there,
  and the numeral marks the generation pass. E and F carry one row each
  because each predicts a single stream: the inverse model produces actions, the forward model observations.
  }
  \label{fig:openwam-masks}
  % \vspace{-0.1in}
\end{figure}

\paragraph{Shared architecture.}
We couple the pretrained video backbone to a separate action
transformer using a Mixture-of-Transformers (MoT) architecture \cite{liang2024mixture}. Video and
action trajectories are encoded into temporally aligned streams and
processed by modality-specific transformer branches. Interleaved
cross-modal attention determines how information flows between them. The
backbone, tokenization, and objective family are shared across
configurations; future cross-modal visibility and generation order are the
principal variables. Figure~\ref{fig:openwam-mot-programs} summarizes the
architecture, and Figure~\ref{fig:openwam-masks} the attention masks of four
interaction programs together with the two local-context dynamics programs.

\paragraph{Interaction programs.}
We study four video--action interaction programs under a shared
Mixture-of-Transformers backbone. \emph{Joint} denoises future video and
action streams concurrently, with bidirectional attention between their
noisy future tokens. \emph{VTA} first predicts future video without access
to future actions, then predicts the action conditioned on the
completed visual trajectory. \emph{ATV} reverses this order: it first
predicts the action, then conditions video prediction on the
completed action trajectory. \emph{Decoupled} predicts both modalities
without future cross-modal attention. All four programs use the same backbone, tokenization, and
training objective; they differ only in generation order and future
cross-modal attention.

\begin{table}[t]
    \centering
    \caption{LIBERO-90 targets used for component-transfer evaluation.}
    % \vspace{-0.1in}
    \label{tab:transfer-tasks}
    \small
    \resizebox{\linewidth}{!}{%
    \begin{tabular}{@{}cll@{}}
        \toprule
        \textbf{Task}
        & \textbf{Target behavior}
        & \textbf{Transfer challenge} \\
        \midrule
        64
        & Stack bowls; place stack in tray
        & Novel skill composition \\
        74
        & Book into caddy's left compartment
        & Retarget familiar manipulation \\
        21
        & Turn on stove; place frying pan on it
        & Object and grasp change \\
        45
        & Same objective as Task 21
        & Additional scene and layout shift \\
        \bottomrule
    \end{tabular}}
    % \vspace{-0.2in}
\end{table}

\begin{figure*}[t]
  \centering
  \includegraphics[width=0.98\linewidth]{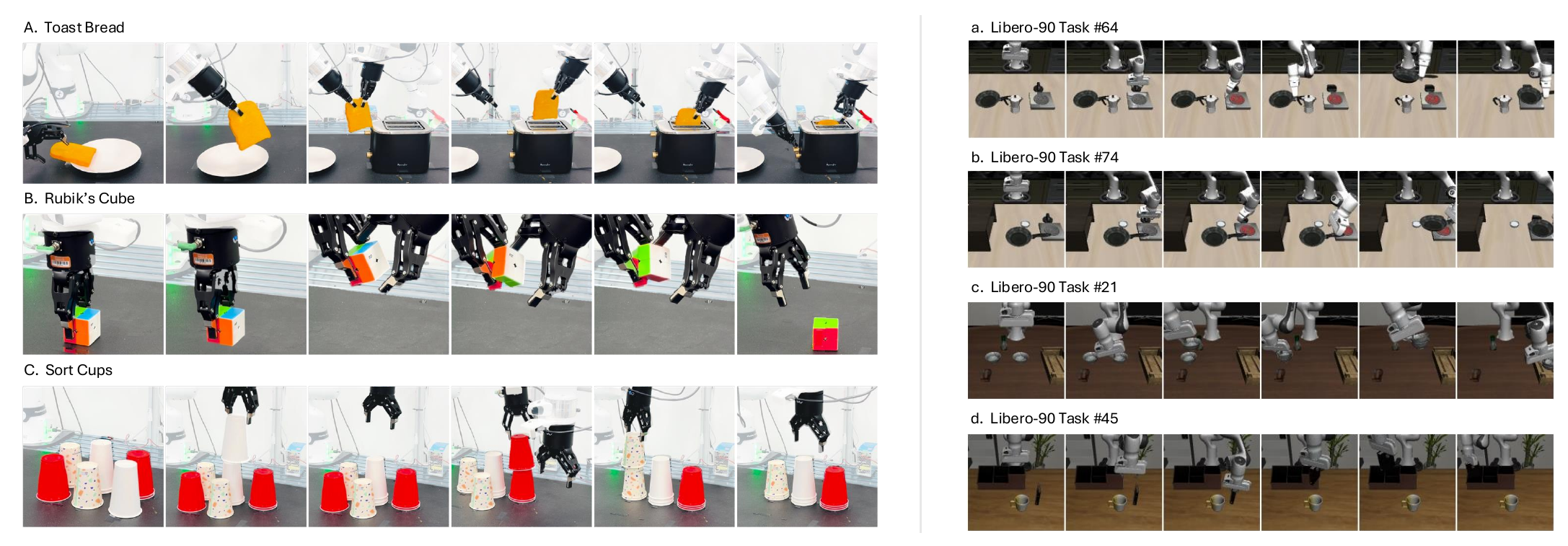}
  % \vspace{-0.2in}
  \caption{Evaluation tasks. \textbf{Left:} the three real-world bimanual
  tasks of Table~\ref{tab:real-world-tasks}, each shown as frames sampled
  along one successful rollout. \textbf{Right:} the four LIBERO-90 targets of Table~\ref{tab:transfer-tasks} used for component transfer, which lie
  outside the LIBERO-Long source-task set.}
  \label{fig:tasks}
  % \vspace{-0.1in}
\end{figure*}

\paragraph{Shared flow-matching objective.}
All programs use the same latent flow-matching~\cite{lipman2022flow} construction. For target
modality $P\in\{\mathrm O,\mathrm A\}$, let $Z_P^\star$ be its clean target
latent. With $\tau\sim\mathcal U[0,1]$ and
$\epsilon_P\sim\mathcal N(0,I)$,
% \begin{equation}
% \begin{aligned}
% Z_{P,\tau}&=(1-\tau)\epsilon_P+\tau Z_P^\star,\\
% u_P&=Z_P^\star-\epsilon_P.
% \end{aligned}
% \label{eq:flow-path}
% \end{equation}
\begin{equation}
Z_{P,\tau}=(1-\tau)\epsilon_P+\tau Z_P^\star,\quad
u_P=Z_P^\star-\epsilon_P.
\label{eq:flow-path}
\end{equation}
An active prediction stage minimizes
\begin{equation}
\mathcal L_P
=
\mathbb E\!\left[
\left\|
v_{\theta,P}
\!\left(
Z_{P,\tau}\mid c_P,\tau
\right)
-u_P
\right\|_2^2
\right],
\label{eq:shared-flow-objective}
\end{equation}
where $c_P$ denotes the history, side information, and cross-modal
trajectory made visible by the interaction program. Programs with
multiple targets sum the corresponding losses. Observation and
action trajectories used as conditioning are processed through their
respective modality encoders.

% During training, the second stage of a sequential program conditions on the
% recorded cross-modal trajectory rather than a sampled first-stage output;
% this conditioning may be stochastically corrupted according to the training
% configuration. At inference, it is replaced by the completed trajectory
% generated by the first stage. The complete token-level visibility masks,
% corruption schedules, temporal alignment, chunk configuration, and sampling
% procedure are provided in the appendix.
During training, a sequential program uses teacher forcing: the second
generation pass conditions on the recorded cross-modal trajectory rather than
on an output sampled by the first pass. At inference, it conditions on the
completed trajectory generated by the first pass. The same interaction program
and temporal alignment are used during training and inference; the source of the conditioning trajectory changes
from recorded to generated. 
% The history-stream visibility settings of the
% examined checkpoints and evaluation launcher are documented in
% Sec.~\ref{sec:supp-compute}.

\subsection{Counterfactual Learning of Local-Context Dynamics Interfaces}
\label{sec:strict-dynamics}

\paragraph{Local-context dynamics interfaces.}
Architectural composition controls communication within a model, but
reusing a separately trained component requires an explicit conditioning
boundary. A task-conditioned generator may use language and nonlocal
history to specify behavior; a reusable local dynamics component instead
relates a supplied transition to its actions, or supplied actions to their
consequences.

We therefore define
\begin{equation}
\begin{aligned}
p_{\mathrm{IDM}}^{\mathrm{local}}
&:=
p(
 A_t^+
\mid
 O_t^+, O_t^0,q_t
),\\
p_{\mathrm{FDM}}^{\mathrm{local}}
&:=
p(
 O_t^+
\mid
 A_t^+, O_t^0,q_t
).
\end{aligned}
\label{eq:strict-modeling-roles}
\end{equation}
% \begin{equation}
% p_{\mathrm{IDM}}^{\mathrm{local}} := p(A_t^+ \mid O_t^+, O_t^0, q_t),
% \qquad
% p_{\mathrm{FDM}}^{\mathrm{local}} := p(O_t^+ \mid A_t^+, O_t^0, q_t).
% \label{eq:strict-modeling-roles}
% \end{equation}
Local-context IDM grounds a supplied visual trajectory into an action sequence,
while local-context FDM predicts the visual consequence of a supplied action
sequence. Neither receives language, past actions, or observations
preceding $t$.

Importantly, conditioning scope and transition supervision are separate
design choices. The local-context restriction specifies which variables the dynamics component
may consume; it does not determine which transitions are used for learning.
Conversely, broader transition supervision can also be applied to a
full-context dynamics model. Sec.~\ref{sec:component-composability}
therefore compares full-context and local-context IDMs under matched
demonstration-only and mixed supervision.

% Local-context conditioning is an input restriction, not an assumption that the
% current observation is universally sufficient to determine local dynamics.
% Formal local-sufficiency conditions and their qualifications are provided
% in the appendix; we evaluate the adequacy of the restricted interface
% empirically through local prediction and composed control.
Local-context conditioning restricts the variables available to the dynamics
component; it does not assume universal local sufficiency. Its adequacy
depends on whether the information needed for local action--outcome inference
is observable from the current observation, proprioceptive state, and supplied
transition. We evaluate this interface through local prediction and composed
control.

\paragraph{Broader local-transition supervision.}
A task trajectory induces a local transition
$z=( O_t^0,q_t, A_t^+, O_t^+)$.
Let $\mathcal D_{\mathrm{task}}^{\mathrm{loc}}$ denote the distribution of
such transitions obtained from demonstrations. Demonstrations provide valid
transitions but contain only the action continuations selected by the
policy.

To broaden local supervision, we restore selected simulator states and
execute alternative future action sequences. For branch $b$, this produces
$z^{(b)}=( O_t^0,q_t, A_t^{+,(b)},
 O_t^{+,(b)})$, where the initial condition is shared and each
future observation trajectory results from executing its corresponding
action sequence. These alternatives include interactions outside the
demonstrated continuation and need not complete the task. The simulator
state is used only to control data collection and is not provided to the
model.

The paired collection protocol controls the starting state, while the
learning objective operates on individual transition records rather than
using an explicit paired-branch loss. Let $\mathcal D_{\mathrm{cf}}$ denote
the resulting counterfactual transitions. Dynamics models are trained from
\begin{equation}
\mathcal D_{\mathrm{dyn}}
=
(1-\eta)\mathcal D_{\mathrm{task}}^{\mathrm{loc}}
+
\eta\mathcal D_{\mathrm{cf}},
\qquad
\eta\in[0,1].
\label{eq:dynamics-training-distribution}
\end{equation}
This supports demonstration-only, counterfactual-only, and mixed
supervision without changing the conditioning interface. The experimental
mixture and collection details are given in Sec.~\ref{sec:experimental-setup}.
% The same local-transition representation also accommodates naturally occurring failed attempts and play in the physical-robot setting.
The same interface can be applied to failed or exploratory physical-robot transitions when such data are collected.

\paragraph{Local dynamics learning.}
Using the flow construction in Eq.~\ref{eq:flow-path}, local-context IDM denoises
the action target while conditioning on the observed future, whereas local-context
FDM denoises the future observation target while conditioning on the
executed action sequence:
\begin{equation}
\begin{aligned}
\mathcal L_{\mathrm{LC\text{-}IDM}}
&=
\mathbb E\!\left[
\left\|
v_{\theta_{\mathrm{ID}}}
(
Z_{\mathrm A,\tau}
\mid
 O^0,q_t, O^+,\tau
)
-u_{\mathrm A}
\right\|_2^2
\right],\\
\mathcal L_{\mathrm{LC\text{-}FDM}}
&=
\mathbb E\!\left[
\left\|
v_{\theta_{\mathrm{FD}}}
(
Z_{\mathrm O,\tau}
\mid
 O^0,q_t, A^+,\tau
)
-u_{\mathrm O}
\right\|_2^2
\right].
\end{aligned}
\label{eq:strict-dynamics-objectives}
\end{equation}
Ground-truth future observations condition IDM training but are not video
denoising targets; analogously, executed actions condition FDM training.

\paragraph{Inference and component composition.}

For video-to-action composition, the video predictor passes future
variational autoencoder (VAE) video latents to the receiving IDM;
transformer hidden states and caches are not transferred across components.
The receiving IDM processes the transferred latents through the same
modality-conditioning pathway used for recorded conditioning.
% Detailed tensor layouts and encoding configuration are given in the appendix.

At inference, a task-conditioned visual predictor can be paired with an
independently trained local-context IDM:
% \begin{equation}
% \begin{aligned}
% \hat{ O}_t^+
% &\sim
% p_\phi^{\mathrm{CVP}}
% (\cdot\mid\mathcal C_t),\\
% \hat{ A}_t^+
% &\sim
% p_{\theta_{\mathrm{ID}}}^{\mathrm{LC\text{-}IDM}}
% (
% \cdot
% \mid
% \hat{ O}_t^+, O_t^0,q_t
% ).
% \end{aligned}
% \label{eq:strict-vta-composition}
% \end{equation}
\begin{equation}
\hat{O}_t^+ \sim p_\phi^{\mathrm{CVP}}(\cdot\mid\mathcal C_t),
\quad
\hat{A}_t^+ \sim p_{\theta_{\mathrm{ID}}}^{\mathrm{LC\text{-}IDM}}
(\cdot\mid \hat{O}_t^+, O_t^0, q_t).
\label{eq:strict-vta-composition}
\end{equation}
Task context determines the desired future through the visual predictor,
while the receiving IDM receives only the generated future trajectory and the
current local inputs.
% Detailed temporal masking is given in the appendix.

The complementary composition pairs an action generator with local-context FDM:
% \begin{equation}
% \begin{aligned}
% \hat{ A}_t^+
% &\sim
% p_\psi^{\mathrm{VLA}}
% (\cdot\mid\mathcal C_t),\\
% \hat{ O}_t^+
% &\sim
% p_{\theta_{\mathrm{FD}}}^{\mathrm{LC\text{-}FDM}}
% (
% \cdot
% \mid
% \hat{ A}_t^+, O_t^0,q_t
% ).
% \end{aligned}
% \label{eq:strict-atv-composition}
% \end{equation}
\begin{equation}
\hat{A}_t^+ \sim p_\psi^{\mathrm{VLA}}(\cdot\mid\mathcal C_t),
\quad
\hat{O}_t^+ \sim p_{\theta_{\mathrm{FD}}}^{\mathrm{LC\text{-}FDM}}
(\cdot\mid \hat{A}_t^+, O_t^0, q_t).
\label{eq:strict-atv-composition}
\end{equation}

The action sequence may equivalently be supplied by an external policy.
These are learned compositions: separately trained components need not
recover factors of one identical learned joint distribution. Their common
trajectory interfaces instead make the components explicitly replaceable,
and Sec.~\ref{sec:component-composability} evaluates when such replacement
succeeds.

\section{Experiments}
\label{sec:experiments}

We evaluate \ours along three dimensions: \emph{native policy performance},
\emph{component composability}, and \emph{local dynamics modeling}. We test
whether the shared robot-video foundation supports different video--action
interaction programs, whether independently trained action components remain
useful when paired with adapted visual predictors, and whether local-context forward
dynamics models action-dependent visual outcomes.

\subsection{Experimental Setup}
\label{sec:experimental-setup}

\paragraph{Robot-video pretraining and simulation.}
We initialize the visual backbone from Wan2.2-5B~\cite{wan2025wan} and further
pretrain it without action supervision on heterogeneous robot and
human-interaction video, including UMI/MV-UMI~\cite{chi2024universal,rayyan2026mv},
AgiBot World~\cite{bu2025agibot}, RoboMIND~\cite{wu2024robomind},
InternData-A1~\cite{tian2026interndata}, RoboCOIN~\cite{wu2025robocoin},
FastUMI~\cite{zhaxizhuoma2025fastumi}, Ego-Exo4D~\cite{grauman2024ego},
and Open X-Embodiment~\cite{o2024open}.
% Dataset composition, preprocessing, sampling, and optimization details are provided in the
% appendix.
Unless otherwise stated, downstream \ours variants use this
common initialization.

The four interaction programs share the same dual-stream
MoT backbone and downstream training recipe; they
differ in generation order and future cross-modal attention. Video is
processed in four-frame latent chunks, each predicting a 16-step action
chunk, with proprioception provided as a per-chunk conditioning signal.
Actions consist of end-effector position, axis-angle rotation, and a
gripper command. We use a two-stage training procedure: causal
robot-video pretraining followed by joint video--action training on the
target suite.

We evaluate Joint, VTA, ATV, and Decoupled on LIBERO-Object, LIBERO-Goal,
LIBERO-Spatial, and LIBERO-Long~\cite{liu2023libero}.
Component transfer uses LIBERO-Long as the downstream source
distribution and four LIBERO-90 targets (Table~\ref{tab:transfer-tasks}) chosen to represent qualitatively different forms of transfer. These targets are outside the LIBERO-Long
source-task set; robot-video pretraining is treated as a separate stage.

\begin{table}[t]
    \centering
    \caption{Real-world bimanual manipulation tasks.}
    % \vspace{-0.05in}
    \label{tab:real-world-tasks}
    \small
    \resizebox{\linewidth}{!}{%
    \begin{tabular}{@{}lll@{}}
        \toprule
        \textbf{Task}
        & \textbf{Objective}
        & \textbf{Main challenge} \\
        \midrule
        Toast Bread
        & Insert bread and activate toaster
        & Deformability and precise insertion \\
        Rubik's Cube
        & Restore final layer of $2\times2$ cube
        & Contact-rich push and rotation \\
        Sort Cups
        & Sort cups according to color
        & Variable ordering and sequencing \\
        \bottomrule
    \end{tabular}}
    % \vspace{-0.2in}
\end{table}

\paragraph{Real-world tasks.}
We evaluate on a bimanual platform comprising two Franka Research 3 arms
with parallel-jaw grippers and three RGB views: two wrist-mounted cameras
and one centered third-person camera. The three task definitions
are listed in Table~\ref{tab:real-world-tasks}. Figure~\ref{fig:tasks} shows the
real-world tasks and the LIBERO-90 transfer targets.

\paragraph{Dynamics supervision.}
Native task models are trained from task demonstrations. For local dynamics,
we additionally construct \textbf{LIBERO-Long-CF} by executing alternative
action continuations from states encountered in LIBERO-Long demonstrations.
Branches sharing an initial state remain in the same train, validation, or
test partition.
IDM and FDM experiments use demonstration-only, counterfactual-only
(CF-only), or mixed supervision. Mixed supervision samples 60\% of
transitions from $\mathcal D_{\mathrm{cf}}$ and 40\% from
$\mathcal D_{\mathrm{task}}^{\mathrm{loc}}$, corresponding to $\eta=0.6$
in Eq.~\ref{eq:dynamics-training-distribution}.

%For physical experiments, the broader dynamics data instead combines successful demonstrations, failed attempts, and free-form play.
% Collection and sampling details are provided in the appendix.

\paragraph{Evaluation.}
Closed-loop task success is the primary policy and composition metric.
For forward dynamics, we report visual prediction error and outcome
identification accuracy. Given $K$ realized futures from the same state,
the metric tests whether the prediction is closest in RGB MSE to its
matching outcome. We report both pairwise identification ($K=2$) and
identification among all 16 action branches ($K=16$).

\begin{table}[!t]
    \centering
    \caption{Closed-loop success (\%) on four LIBERO suites.
Baselines are previously reported results.}
    % \vspace{-0.1in}
    \label{tab:libero-broad}
    \scriptsize
    \setlength{\tabcolsep}{5pt}
    \begin{tabular}{@{}lccccc@{}}
        \toprule
        \textbf{Method}
        & \textbf{Object}
        & \textbf{Goal}
        & \textbf{Spatial}
        & \textbf{Long}
        & \textbf{Mean} \\
        \midrule
        OpenVLA~\cite{kim2024openvla}
        & 88.4 & 79.2 & 84.7
        & 53.7 & 76.5 \\
        OpenVLA-OFT~\cite{kim2025fine}
        & 98.4 & 97.9 & 97.6
        & 94.5 & 97.1 \\
        $\pi_0$~\cite{black2024pi_0}
        & 98.8 & 95.8 & 96.8
        & 85.2 & 94.1 \\
        $\pi_{0.5}$~\cite{intelligence2025pi_}
        & 98.2 & 98.0 & 98.8
        & 92.4 & 96.9 \\
        GR00T-N1~\cite{bjorck2025gr00t}
        & 97.6 & 93.0 & 94.4
        & 90.6 & 93.9 \\
        Motus~\cite{bi2026motus}
        & 99.8 & 96.6 & 96.8
        & 97.6 & 97.7 \\
        Fast-WAM~\cite{yuan2026fast}
        & 100.0 & 97.0 & 98.2
        & 95.2 & 97.6 \\
        LingBot-VA~\cite{li2026causal}
        & 99.6 & 97.2 & 98.5
        & 98.5 & 98.5 \\
        \midrule
        \textbf{\ours-VTA}
        & $\text{99.4}_{\pm\text{0.3}}$
        & $\text{98.4}_{\pm\text{0.5}}$
        & $\text{98.6}_{\pm\text{0.2}}$
        & $\text{97.8}_{\pm\text{0.4}}$
        & 98.6 \\

        \textbf{\ours-ATV}
        & $\text{98.0}_{\pm\text{0.4}}$
        & $\text{97.2}_{\pm\text{0.2}}$
        & $\text{96.6}_{\pm\text{0.6}}$
        & $\text{95.4}_{\pm\text{0.3}}$
        & 96.8 \\

        \textbf{\ours-Joint}
        & $\text{98.2}_{\pm\text{0.2}}$
        & $\text{97.8}_{\pm\text{0.4}}$
        & $\text{97.6}_{\pm\text{0.3}}$
        & $\text{96.6}_{\pm\text{0.5}}$
        & 97.6 \\

        \textbf{\ours-Decoupled}
        & $\text{99.0}_{\pm\text{0.3}}$
        & $\text{98.0}_{\pm\text{0.2}}$
        & $\text{97.8}_{\pm\text{0.5}}$
        & $\text{97.0}_{\pm\text{0.4}}$
        & 98.0 \\
        \bottomrule
    \end{tabular}
    % \vspace{-0.1in}
\end{table}

\begin{table}[!t]
    \centering
    \caption{
    Closed-loop success (\%) on real-world bimanual manipulation tasks.
    }
    % \vspace{-0.1in}
    \label{tab:real-world-native}
    \small
    \begin{tabular}{@{}lcccc@{}}
        \toprule
        \textbf{Method}
        & \textbf{Toast}
        & \textbf{Cube}
        & \textbf{Cups}
        & \textbf{Mean} \\
        \midrule
        \ours-VTA
        & 92.0 & 90.0 & 94.4 & 92.1 \\
        \ours-Joint
        & 90.0 & 94.0 & 91.7 & 91.9 \\
        \bottomrule
    \end{tabular}
    \vspace{-0.2in}
\end{table}

\begin{table*}[t]
    \centering
    \caption{
    Frozen inverse dynamics transfer to four LIBERO-90 targets.
    Target rows use task-tuned video predictors while source-trained IDMs
    remain frozen. Mean is averaged over the four target tasks.
    }
    % \vspace{-0.1in}
    \label{tab:component-transfer}
    \small
    \setlength{\tabcolsep}{5pt}
    \begin{tabular}{@{}llccccc@{}}
        \toprule
        \textbf{Video / policy}
        & \textbf{Action component (frozen)}
        & \shortstack{\textbf{64}\\Composition}
        & \shortstack{\textbf{74}\\Retarget}
        & \shortstack{\textbf{21}\\Object/grasp}
        & \shortstack{\textbf{45}\\Scene shift}
        & \shortstack{\textbf{Transfer}\\\textbf{mean}} \\
        \midrule

        \multicolumn{2}{l}{
            \textbf{Task-tuned VTA (finetuned policy reference)}
        }
        & 94\%
        & 86\%
        & 92\%
        & 100\%
        & 93.0\% \\

        \midrule

        Task-tuned video
        & Full-context IDM, demo-only
        & 88\%
        & 90\%
        & 0\%
        & 0\%
        & 44.5\% \\

        Task-tuned video
        & Full-context IDM, mixed
        & 54\%
        & 92\%
        & 4\%
        & 38\%
        & 47.0\% \\

        \midrule

        Task-tuned video
        & Local-context IDM, demo-only
        & 36\%
        & 50\%
        & 0\%
        & 0\%
        & 21.5\% \\

        Task-tuned video
        & Local-context IDM, CF-only
        & 80\%
        & 92\%
        & 90\%
        & 76\%
        & \textbf{84.5\%} \\

        Task-tuned video
        & Local-context IDM, mixed
        & 74\%
        & 86\%
        & 88\%
        & 88\%
        & \textbf{84.0\%} \\

        \midrule

        \multicolumn{2}{l}{
            \textbf{Original VTA (unadapted policy reference)}
        }
        & 42\%
        & 0\%
        & 0\%
        & 0\%
        & 10.5\% \\

        \bottomrule
    \end{tabular}
    \vspace{-0.1in}
\end{table*}

% ---------------------------------------------------------
% Source-task composition on LIBERO-Long
% ---------------------------------------------------------
\begin{table}[!t]
    \centering
    \caption{
    Source-task composition on LIBERO-Long, all action components evaluated with LIBERO-Long video
    predictor.
    }
    % \vspace{-0.1in}
    \label{tab:component-source}
    \footnotesize
    \begin{tabular}{@{}llc@{}}
        \toprule
        \textbf{Action component}
        & \textbf{Supervision}
        & \textbf{Success (\%)} \\
        \midrule

        Full-context IDM
        & Demo-only (Native VTA)
        & 97.8 \\

        Full-context IDM
        & Mixed
        & 92.2 \\

        \midrule

        Local-context IDM
        & Demo-only
        & 25.8 \\

        Local-context IDM
        & CF-only
        & 90.6 \\

        Local-context IDM
        & Mixed
        & \textbf{94.4} \\

        \bottomrule
    \end{tabular}
    % \vspace{-0.1in}
\end{table}

\begin{table}[t]
    \centering
    \caption{
    Local-context FDM prediction on counterfactual transitions.
    MSE values are reported in units of $10^{-3}$.
    Outcome accuracy measures identification of the matching realized future
    by RGB MSE among $K$ same-state outcomes.
    }
    \label{tab:fdm}
    \small
    \begin{tabular}{@{}lcccc@{}}
        \toprule
        \textbf{Supervision}
        & \textbf{RGB MSE $\downarrow$}
        & \textbf{Acc.} $K{=}2$ $\uparrow$
        & \textbf{Acc.} $K{=}16$ $\uparrow$ \\
        \midrule
        Demo-only
        & 14.35
        & 68.3
        & 21.1 \\
        CF-only
        & \textbf{9.40}
        & \textbf{93.6}
        & \textbf{71.3} \\
        Mixed
        & 9.62
        & 91.8
        & 67.7 \\
        \bottomrule
    \end{tabular}
    \vspace{-0.2in}
\end{table}

\subsection{Native Policy Performance}
\label{sec:native-policy-performance}

\paragraph{Broad LIBERO evaluation.}
% We first test whether the shared foundation supports strong native policies
% across different generation orders, and report the average success rate over three random seeds in Table~\ref{tab:libero-broad}.
We compare the four interaction programs across the LIBERO suites and
report the average success rate over three random seeds in
Table~\ref{tab:libero-broad}.
The principal \ours configurations achieve strong performance across all
four suites, establishing that the shared foundation supports multiple
world--action generation orders.

\paragraph{Backbone initialization ablation.}
Holding the VTA architecture, task supervision, and downstream training
procedure fixed, the original Wan2.2-5B initialization reaches 68.4\%
success on LIBERO-Long, while our robot-video-pretrained causal backbone
reaches 97.8\% (+29.4 points). This comparison measures the combined effect
of robot-video pretraining and causal adaptation, rather than attributing the
gain to either factor individually.

\paragraph{Real-world native policies.}

% The physical evaluation tests the same framework under multi-view bimanual,
% deformable, contact-rich, and multi-object manipulation.
Table~\ref{tab:real-world-native} reports closed-loop success for VTA and
Joint on the three bimanual tasks of Table~\ref{tab:real-world-tasks}, which
test the same framework under multi-view, deformable, contact-rich,
and multi-object manipulation.

\subsection{Component Composability}
\label{sec:component-composability}

% ---------------------------------------------------------
% Target-task transfer on LIBERO-90
% ---------------------------------------------------------

We next study whether action-grounding components remain useful when the visual predictor changes. Full-context and local-context IDMs are evaluated under transition-supervision regimes, separating the effects of broader local-transition data from the restriction of task-level context.

\paragraph{Frozen action-component reuse.}
For each LIBERO-90 target, the visual producer is taken from a task-tuned
VTA model while source-trained action components remain frozen. Thus, the
experiment measures reuse of an action component after visual-task
adaptation, rather than target-task learning without action supervision.
All reused executors are evaluated with the same target predictor and
rollout protocol. Table~\ref{tab:component-source} reports source-task
composition on LIBERO-Long.

Conditioning scope and supervision interact strongly
(Tables~\ref{tab:component-source} and~\ref{tab:component-transfer}).
Mixed supervision raises mean target success from 21.5\% to 84.0\%
for local-context IDMs, compared with 44.5\% to 47.0\% for
full-context IDMs. Under matched mixed supervision, local conditioning
improves target success by 37.0 percentage points while retaining
94.4\% source-task success.
% Neither local conditioning nor broader
% supervision alone matches their combination in this evaluation.
Counterfactual-only and mixed supervision both substantially improve
local-context IDM transfer, achieving comparable mean success of 84.5\%
and 84.0\%, respectively, across the four LIBERO-90 targets. These
results show that broader transition coverage supports the reuse of
local-context dynamics components after visual task adaptation.

The per-task results also reveal uneven transfer. The demonstration-only
full-context IDM performs well on Tasks 64 and 74 but obtains 0\%
success on Tasks 21 and 45, whereas CF-only and mixed local-context
IDMs remain effective across all four targets. This contrast highlights
their more consistent reuse across the evaluated task changes.

\subsection{Action-Conditioned Forward Dynamics}

We evaluate local-context FDMs on 2,560 counterfactual futures from
160 contexts across ten LIBERO tasks, with 16 action branches per context.
Models share the same initial observations, actions, and realized futures.
In addition to RGB prediction error, we measure whether the predicted future
preserves the identity of its conditioning action through outcome
identification at $K=2$ and $K=16$.

As shown in Table~\ref{tab:fdm}, CF-only and mixed supervision reduce RGB MSE
by 34.5\% and 33.0\%, respectively, relative to demonstration-only training.
CF-only supervision raises outcome accuracy from 68.3\% to 93.6\% for
$K=2$, and from 21.1\% to 71.3\% for $K=16$. Mixed supervision shows a
similar improvement, reaching 91.8\% and 67.7\%, respectively. These results
show that broader transition supervision improves both visual prediction
accuracy and the ability to distinguish action-dependent futures.
\section{Discussion and Conclusion}
\label{sec:discussion}

We presented \ours, an open framework and pretrained foundation spanning
video prediction, robot control, and local dynamics modeling. Its shared
causal backbone supports joint, sequential, and decoupled video--action
generation within a common training and evaluation stack. Strong performance
across four LIBERO suites and real-world bimanual tasks demonstrates the
framework's control capabilities, while the initialization ablation
establishes the substantial benefit of robot-video pretraining. \ours
thus supports multiple effective modeling configurations rather than a
single specialized policy.

% Beyond native control, the framework enables independently trained
% inverse and forward dynamics components. Our controlled simulation study
% shows that local conditioning and counterfactual supervision jointly
% support substantially stronger frozen-executor reuse after visual task
% adaptation than either choice alone, while retaining strong source-task
% performance. Counterfactual supervision also improves prediction of
% alternative action outcomes, connecting broader interaction data to useful
% forward modeling.
Beyond native control, the framework supports independently trained
inverse and forward dynamics components. In the transfer study,
counterfactual-only and mixed supervision both substantially improve the
reuse of local-context IDMs after visual task adaptation, while retaining
strong source-task performance. Counterfactual FDM supervision also improves both visual prediction accuracy and
identification of action-dependent outcomes.

Several questions remain open. Composability is established primarily in
simulation; demonstrating comparable reuse on real robots will require more
data-efficient physical dynamics supervision. Our target-task experiments
adapt the video predictor and therefore do not establish zero-shot transfer of
the complete video--action model. Appendix~E shows that policy, IDM, and FDM
objectives can be trained jointly in one checkpoint, but we have not yet shown
that a single multi-objective model can match the strongest specialized
\ours variants across all modes. The present FDM study also focuses on local,
short-horizon prediction, leaving long-horizon planning and physical deployment
as natural extensions.

\ours brings these capabilities into one open foundation: strong
policies, configurable prediction--action interaction, and independently
reusable dynamics. It provides both capable world--action models and a
common basis for building, comparing, and composing the next generation.

\FloatBarrier

{
    \small
    \bibliographystyle{ieeenat_fullname}
    \bibliography{main}
}

\clearpage
\setcounter{page}{1}

\clearpage
% \onecolumn          % switch to full-width *before* supplementary

\appendix
% \suppressfloats[t]

\begin{center}
    \Large\bfseries Appendix
\end{center}
\vspace{1em}

\section{Implementation Details}
\label{sec:supp-impl}

This section provides implementation and evaluation details for the four video--action interaction programs studied in the main paper, including model architecture, downstream training, generation, and closed-loop evaluation.

\subsection{Model Architecture and Representation}
\label{sec:supp-architecture}

\paragraph{Dual-stream backbone.}
We use a 30-layer dual-stream Mixture-of-Transformers (MoT) architecture with modality-specific video and action experts.
The video expert has a hidden dimension of 3072, 24 attention heads with a head dimension of 128, and an FFN dimension of 14336. The action expert has a hidden dimension of 2048 and an FFN dimension of 8192, while retaining the same attention configuration of 24 heads with a head dimension of 128. Consequently, its query, key, and value projections each map the 2048-dimensional hidden state to 3072 dimensions, and the attention output is projected back to 2048 dimensions. 
Cross-modal information exchange is controlled by the future cross-modal attention pattern associated with each interaction program.
Task text conditions both streams through a separate cross-attention layer.

\paragraph{Visual observations.}
Each observation contains an agent-view image and a wrist-camera
image, both at $128\times128$ resolution. 
The two views are encoded independently into \(8\times8\times48\) VAE latents and concatenated along the spatial width dimension to form an \(8\times16\times48\) latent canvas.
Using a \(1\times2\times2\) temporal-spatial patch size, each latent frame contains 32 video tokens.
The video VAE temporally compresses every four environment steps into one latent video frame. Each generation chunk spans four future latent frames, corresponding to 16 environment steps.

\paragraph{Actions and proprioception.}
The policy predicts a 16-step action chunk aligned with the same temporal horizon, with four action steps corresponding to each latent video frame. Each action has seven dimensions: three for end-effector position, three for axis-angle rotation, and one for the gripper command. Each chunk is conditioned on an eight-dimensional proprioceptive state comprising the 3D end-effector position, a 3D axis-angle orientation converted from the simulator quaternion, and two gripper joint positions. The proprioceptive state is projected to the corresponding hidden dimension of each stream and added to its hidden representation rather than appended as a sequence token.

\subsection{Downstream Training}
\label{sec:supp-training}

The four interaction programs start from the same robot-video-pretrained causal backbone. The action expert is initialized from width-adapted copies of the corresponding pretrained video-expert layers, while action-specific input/output and conditioning layers are initialized separately. The video and action experts are then jointly optimized during downstream training. All programs use the same data, optimization settings, and flow-matching objective; they differ only in generation order and future cross-modal attention.

For the downstream runs examined here, training uses AdamW with
$\beta_1=0.9$, $\beta_2=0.95$, weight decay 0.1, a learning rate of
$10^{-5}$, with 10 warmup steps followed by a constant learning-rate schedule.
We use a per-device batch size of 1 and accumulate gradients over 10 steps, yielding an effective batch size of 10. Gradient norms
are clipped at 2.0, and training uses bf16 mixed precision.
We train all downstream models for 10,000 optimization steps. 
% All downstream models are trained on a single NVIDIA B200 GPU.

% \paragraph{Flow-matching objective.}
% Both streams are trained with a rectified-flow objective. Following
% the time convention used in the main text, a clean target $x_0$ and
% Gaussian noise $\epsilon$ define the interpolated input
% \begin{equation}
% x_{\tau}=(1-\tau)\epsilon+\tau x_0,
% \qquad
% u=x_0-\epsilon.
% \end{equation}
% Training times are sampled uniformly before the stream-specific
% noise schedules are applied. The video and action streams use
% separate noise schedules, with shift factors of 5.0 and 1.0,
% respectively. The flow-matching loss is applied to future video
% latents and valid action dimensions, with each modality normalized
% by its corresponding mask. Both losses use timestep-dependent
% weighting.
\paragraph{Flow-matching objective.}
Both streams are trained with the rectified-flow objective in
Eq.~(4). For implementation, we parameterize the noise level as
$\sigma = 1-\tau$, such that $\sigma=0$ denotes clean data and
$\sigma=1$ denotes pure noise. We uniformly sample from a
1,000-level base noise grid and apply the modality-specific shift
\begin{equation}
S_s(\sigma)
=
\frac{s\sigma}{1+(s-1)\sigma},
\qquad
s_V=5.0,\quad s_A=1.0,
\end{equation}
for video and action, respectively.

We additionally apply timestep-dependent loss weighting,
\begin{equation}
w_P(\sigma)
\propto
\exp\!\left[-2\left(\sigma-\tfrac{1}{2}\right)^2\right]
-
e^{-1/2},
\end{equation}
where the weights are normalized to have unit mean over the
corresponding timestep grid. This emphasizes intermediate noise
levels while assigning zero weight to the pure-noise endpoint.

The flow-matching loss is applied only to predicted future video
latents and valid action entries. Video losses are normalized over
the supervised latent elements, while action losses are averaged
over the action horizon after masking invalid action dimensions.
The two modality losses are combined with equal weight,
\begin{equation}
\mathcal{L}
=
\mathcal{L}_{V}
+
\mathcal{L}_{A}.
\end{equation}

\subsection{Video--Action Interaction Programs}
\label{sec:supp-programs}

The four programs differ in how predicted video and action
trajectories interact within a generation chunk. Each program
also receives the observed history and task instruction.

\paragraph{Joint.}
Video and action trajectories are denoised concurrently. During
generation, each stream can attend to the other stream's evolving
prediction, allowing bidirectional interaction before either
trajectory is complete.

\paragraph{Video-to-action (VTA).}
The model first generates the future video trajectory without
conditioning on future actions. It then generates the action
trajectory conditioned on the generated video trajectory.

\paragraph{Action-to-video (ATV).}
The model first generates the future action trajectory without
conditioning on future video. It then generates the video trajectory
conditioned on the generated action trajectory.

\paragraph{Decoupled.}
Video and action trajectories are generated concurrently, without
cross-modal interaction between their future predictions. Both
streams can still use the observed history and task instruction.

During training, the second stage of VTA and ATV conditions on the ground-truth trajectory of the first modality. 
At inference time, this ground-truth trajectory is replaced by the trajectory generated in the first stage. 
Joint and Decoupled jointly denoise both modalities in a single generation stage, whereas VTA and ATV use two sequential generation stages.

% \subsection{Inference and Closed-Loop Evaluation}
% \label{sec:supp-evaluation}

% \paragraph{Sampling.}
% Inference uses 20 denoising steps for video and 20 for actions with
% a UniPC solver. Video generation uses classifier-free guidance at
% scale 5.0; action generation uses a guidance scale of 1.0. Key--value
% caching is enabled. The inference attention window is 30 latent
% frames.

% \paragraph{Action execution.}
% By default, the policy executes the full predicted action chunk
% before replanning. Episodes are limited to 800 environment steps
% or 50 generated chunks. The rollout setup uses one model-observation
% frame at startup and five environment initialization steps.

% \paragraph{Evaluation protocol.}
% For each evaluated checkpoint, the recorded closed-loop protocol
% covers 10 tasks with 50 episodes per task. The episode index
% determines the environment seed, so programs evaluated under this
% protocol encounter the same set of initial conditions. Success is
% measured from the resulting environment rollouts.
\subsection{Inference and Closed-Loop Evaluation}
\label{sec:supp-evaluation}

\paragraph{Sampling.}
We use 20 denoising steps for each modality, with the same
modality-specific timestep schedules as in training. Video generation
uses classifier-free guidance with scale 5.0, while no classifier-free
guidance is applied to actions. Key--value caching is enabled during
inference. The history window spans 15 generation chunks, corresponding
to 60 latent video frames and the temporally aligned action history.

% \paragraph{Action execution.}
% The policy normally executes the full 16-step action chunk before
% replanning. For the first chunk, only 12 actions are executed because
% the first latent-frame group overlaps the startup observation. After
% resetting to the episode's initial state, the simulator is advanced for
% five zero-action steps to settle, and the policy is initialized from the
% final observation. Episodes terminate upon task success, after 800
% environment steps (including the five initialization steps), or after
% 50 generated chunks, whichever occurs first.
\paragraph{Action execution.}
After resetting to the episode's initial state, the simulator is advanced for
five zero-action steps to settle, and the policy is initialized from the final
observation. The first generation chunk conditions on this single observation;
the same startup convention is used during training, so no additional history
frames are required before the first prediction. The policy executes each
16-step action chunk before replanning. Episodes terminate upon task success,
after 800 environment steps (including the five initialization steps), or after
50 action chunks, whichever comes first.

% \paragraph{Evaluation protocol.}
% We evaluate each checkpoint on 10 tasks with 50 closed-loop episodes per
% task, for 500 episodes in total. Within each task, all interaction
% programs use the same 50 environment seeds and therefore share the same
% set of initial conditions. An episode is counted as successful if the
% simulator's task-success predicate is satisfied at any environment step,
% at which point the rollout terminates. We report the mean success rate
% over the 500 episodes.
\paragraph{Evaluation protocol.}
For each interaction program, we evaluate three independently trained
seeds. Each checkpoint is evaluated on 10 tasks with 50 closed-loop
episodes per task, for 500 episodes per seed. Within each task, all
interaction programs use the same 50 environment seeds and therefore
share the same set of initial conditions. An episode is counted as
successful if the simulator's task-success predicate is satisfied at
any environment step, at which point the rollout terminates. We report
the mean and standard deviation of the success rate across the three seeds.

\subsection{Real Robot Setup}

\begin{figure}[t]
    \centering
    \includegraphics[width=0.98\linewidth]{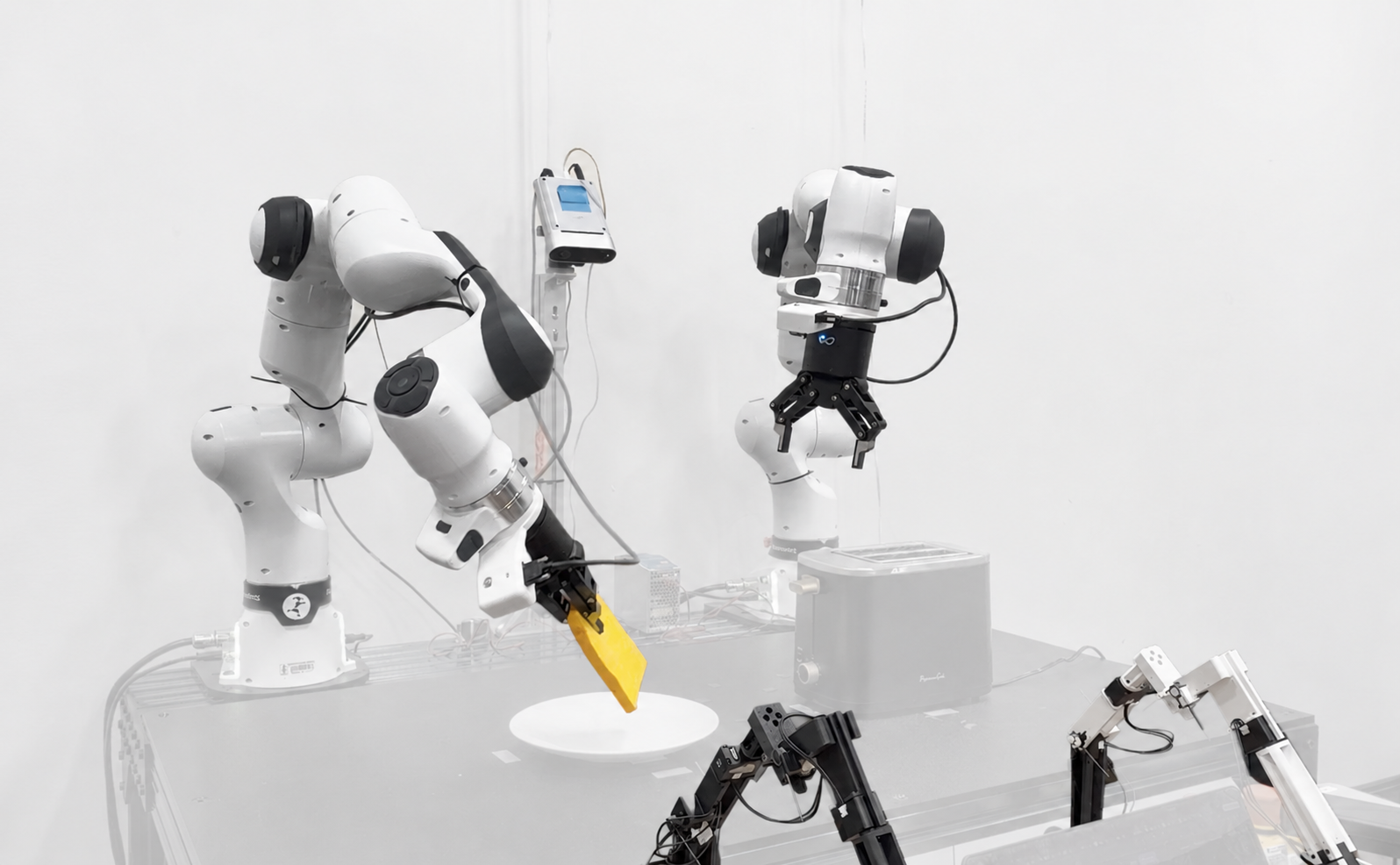}
    \caption{Real-world bimanual setup with two Franka FR3 arms, Flexiv Grav grippers, bimanual GELLO teleoperation~\cite{wu2024gello}, two wrist-mounted RealSense D405 cameras, and a central Orbbec Femto Mega camera.}
\label{fig:real_robot_setup}
\end{figure}

The real-robot setup, as shown in Figure~\ref{fig:real_robot_setup}, consists of two Franka FR3 arms with Flexiv Grav grippers. We collect demonstrations through bimanual GELLO teleoperation. Two wrist-mounted RealSense D405 cameras record at 1280 × 720 pixels, and a central Orbbec Femto Mega records at 1920 × 1080 pixels, all at 30 fps. Joint positions, end-effector poses, gripper widths, and control commands are recorded with timestamps at 30 Hz. In the chunk-relative formulation, each arm’s target poses are expressed relative to its observed pose at the start of the chunk, while gripper widths remain absolute. Each action contains 20 values: three position coordinates, six values representing orientation, and one gripper width for each arm.

For the real-robot experiments, we collect 200 demonstrations each for
Toast Bread and Rubik's Cube, and 180 demonstrations for Sort Cups.
Each demonstration is 30\,s long. We evaluate the learned
policies over 50 closed-loop episodes for Toast Bread and Rubik's Cube,
and 36 episodes for Sort Cups.

\begin{table*}[t]
    \centering
    \caption{
    Data used for causal robot-video pretraining. 
    }
    \label{tab:pretraining-corpus}
    \small
    \begin{tabular}{@{}lrrl@{}}
        \toprule
        \textbf{Dataset}
        & \textbf{Trajectories / takes}
        & \textbf{Hours}
        & \textbf{Source type} \\
        \midrule

        Open X-Embodiment (49 manipulation datasets)
        & 1,300,749
        & 1,911.29
        & Multi-embodiment robot data \\

        AgiBot World Beta
        & 1,003,672
        & 2,976.4
        & Real robot manipulation \\

        Ego-Exo4D v2
        & 5,035
        & 221.26
        & Human interaction \\

        InternData-A1
        & 637,498
        & 7,433.91
        & Synthetic robot manipulation \\

        RoboCOIN
        & 183,157
        & 1,306.83
        & Real bimanual manipulation \\

        RoboMIND
        & 107,877
        & 305.5
        & Multi-embodiment robot data \\

        FastUMI-100K
        & 92,823
        & 461.77
        & Human-guided manipulation \\

        UMI family
        & 5,430
        & 24.60
        & Human-guided manipulation \\

        \bottomrule
    \end{tabular}
\end{table*}

\section{Robot-Video Pretraining}
\label{app:video-pretraining}

We adapt the Wan2.2-5B video backbone to robot-centric motion and physical
interaction before introducing action prediction. The pretraining corpus
combines robot demonstrations, human manipulation video, and synthetic robot
trajectories across a broad range of embodiments, viewpoints, objects, and
interaction patterns. No action labels are used during this stage.

\subsection{Pretraining Corpus}
\label{app:pretraining-data}

Our robot-video pretraining corpus combines large-scale real-robot manipulation,
human interaction video, and synthetic robot trajectories. Table~\ref{tab:pretraining-corpus}
summarizes the source data before training-window sampling. The UMI-family entry
aggregates UMI, DexUMI~\cite{xu2025dexumi}, UMI on Legs~\cite{ha2024umilegs}, and MV-UMI.

The corpus contains approximately 3.34 million trajectory or take units and
14.64k nominal hours, spanning heterogeneous robot embodiments, collection
interfaces, viewpoints, human-object interactions, and synthetic manipulation
at scale, providing the visual and interaction prior used by all downstream
\ours models.

\subsection{Causal Adaptation}
\label{app:pretraining-causal}

Pretraining starts from the Wan2.2-5B checkpoint. We retain the pretrained
video representation and optimize the model on the corpus above using the
chunk-causal prediction objective described in Sec.~\ref{sec:method}. Future
chunks may attend to the current observation, preceding observations, and
previously generated chunks, but not to later chunks. Tokens within each
chunk are denoised jointly.

This stage uses video alone. Robot actions, proprioception, and task-specific
policy targets are not supplied to the model. The resulting checkpoint serves
both as a standalone causal video predictor and as the common visual
initialization for the downstream \ours interaction programs.

\subsection{Training Compute}
\label{app:pretraining-compute}

Causal robot-video pretraining is performed on 32 NVIDIA B200 GPUs for
14 days. All downstream \ours variants in the main experiments start from
the resulting checkpoint unless otherwise specified.

\section{LIBERO-Long-CF: Counterfactual Dynamics Data}
\label{app:libero-cf}

LIBERO demonstrations contain task-directed behavior and cover only a small
fraction of the local state--action--outcome distribution. From each visited
state, they provide essentially one successful action continuation. This is
well suited to imitation learning but provides limited supervision for inverse
and forward dynamics, which must model how alternative actions correspond to
alternative physical outcomes.

We construct \textbf{LIBERO-Long-CF} by branching from task-relevant states in
the ten LIBERO-Long tasks and executing modified future action sequences.
The dataset contains 32,000 counterfactual segments, balanced across tasks.
Each segment contains 128 future controls and 129 synchronized two-view
observations. The original tasks, assets, and simulator physics are preserved.

\subsection{Scale and Intervention Coverage}
\label{app:libero-cf-scale}

LIBERO-Long-CF contains 4.10 million control records, compared with 138,090
controls in the 500 source demonstrations, increasing local transition
supervision by approximately $29.7\times$. At 20\,Hz, these correspond to
56.9 and 1.92 control-equivalent hours, respectively.

\begin{table}[t]
    \centering
    \caption{
    Scale of LIBERO-Long-CF and the source LIBERO-Long demonstration corpus.
    }
    \label{tab:libero-cf-scale}
    \small
    \begin{tabular}{@{}lcc@{}}
        \toprule
        \textbf{Quantity}
        & \textbf{Demonstrations}
        & \textbf{LIBERO-Long-CF} \\
        \midrule
        Tasks
        & 10
        & 10 \\
        Stored sequences
        & 500
        & 32,000 \\
        Sequences / task
        & 50
        & 3,200 \\
        Controls / sequence
        & 276.2 mean
        & 128 \\
        Total controls
        & 138,090
        & 4,096,000 \\
        Duration
        & 1.92 h
        & 56.9 h \\
        \bottomrule
    \end{tabular}
\end{table}

The branch generator modifies motion magnitude, direction, temporal structure,
individual action axes, stochastic arm motion, and gripper behavior.
Table~\ref{tab:libero-cf-composition} summarizes the intervention distribution.

\begin{table}[t]
    \centering
    \caption{
    Intervention composition of LIBERO-Long-CF. Categories denote generation
    recipes and may induce overlapping physical effects.
    }
    \label{tab:libero-cf-composition}
    \small
    \begin{tabular}{@{}lr@{}}
        \toprule
        \textbf{Intervention family}
        & \textbf{Fraction (\%)} \\
        \midrule
        Stop / rescale arm motion
        & 9.4 \\
        Reverse / redirect translation
        & 6.3 \\
        Axis biases and pulses
        & 12.5 \\
        Dedicated yaw perturbation
        & 3.1 \\
        Noise / randomized arm controls
        & 12.5 \\
        Dedicated gripper interventions
        & 31.3 \\
        Random-duration arm / gripper interventions
        & 25.0 \\
        \bottomrule
    \end{tabular}
\end{table}

Interventions include stopped and rescaled motion, reversed or redirected
translation, translational and rotational pulses, structured control noise,
piecewise-random commands, and changes to gripper state and timing. All
commands are within LIBERO's normalized controller range.

\subsection{State and Interaction Coverage}
\label{app:libero-cf-coverage}

Within the dataset, 75\% of segments begin from demonstration-prefix states.
The remaining 25\% begin after an additional executed perturbation, extending
the starting-state distribution beyond the demonstrated trajectories.

Among characterized perturbed starts, the end-effector state is on average
3.34\,cm from the nearest point on the corresponding demonstrated path
(median 1.90\,cm). Under thresholds of 1\,cm translation, $5^\circ$ rotation,
or 5\% articulated-joint travel, 61.6\% also contain object configurations
outside the corresponding demonstrated configurations.

The corpus contains substantial physical interaction. As shown in
Table~\ref{tab:libero-cf-interaction}, 90.4\% of characterized segments contain
gripper contact with an object or fixture, 50.0\% contain a detected grasp,
and 72.3\% alter object configuration relative to the same-start reference
continuation.

\begin{table}[t]
    \centering
    \caption{
    Physical interaction statistics for LIBERO-Long-CF. Categories overlap.
    }
    \label{tab:libero-cf-interaction}
    \small
    \begin{tabular}{@{}lr@{}}
        \toprule
        \textbf{Measured interaction}
        & \textbf{Rate (\%)} \\
        \midrule
        Gripper--object / fixture contact
        & 90.4 \\
        Detected grasp
        & 50.0 \\
        Object-configuration effect
        & 72.3 \\
        \bottomrule
    \end{tabular}
\end{table}

\subsection{Representation for Local Dynamics Learning}
\label{app:libero-cf-representation}

Each segment contains two synchronized $128\times128$ RGB streams,
128 seven-dimensional delta-OSC controls, and 129 proprioceptive states.
Actions contain three translation channels, three rotation channels, and one
gripper command.

Both camera streams are encoded with the \ours video VAE. The encoded
sequence contains one boundary observation and 32 future latent frames, with
four controls aligned to each future latent frame.

Local-context IDM conditions on the boundary observation, proprioception, and
future visual trajectory to predict the action sequence. Local-context FDM
conditions on the boundary observation, proprioception, and action sequence to
predict the future visual trajectory. Both objectives exclude task language
and pre-start history.

\section{Ablations}
\subsection{Effect of Video-Backbone Initialization}
\label{app:init-ablation}
We further examine whether the benefit of robot-video pretraining is specific to
VTA or persists across interaction programs. Under otherwise matched downstream
training, we initialize the video backbone either randomly, from the original
Wan2.2 checkpoint, or from our robot-video-pretrained causal checkpoint.
Table~\ref{tab:init-program-ablation} shows a consistent ordering for both VTA
and Joint. Relative to Wan2.2 initialization, robot-video pretraining improves
success by 29.4 percentage points for VTA and 34.4 points for Joint, reaching
97.8\% and 96.6\%, respectively. Random initialization performs substantially
worse for both programs. These results indicate that the benefit of the
robot-video-pretrained backbone transfers across distinct video--action interaction
structures.

\begin{table}[t]
    \centering
    \caption{
    Effect of video-backbone initialization on closed-loop LIBERO-Long success.
    Downstream architecture and training are held fixed within each interaction
    program.
    }
    \label{tab:init-program-ablation}
    \small
    \begin{tabular}{@{}lcc@{}}
        \toprule
        \textbf{Video initialization}
        & \textbf{VTA (\%)}
        & \textbf{Joint (\%)} \\
        \midrule
        Random initialization
        & 20.0
        & 27.2 \\
        Original Wan2.2
        & 68.4
        & 62.2 \\
        Robot-video pretrained
        & \textbf{97.8}
        & \textbf{96.6} \\
        \bottomrule
    \end{tabular}
\end{table}

\begin{table}[t]
    \centering
    \caption{
    Effect of the Mixture-of-Transformers architecture on closed-loop
    LIBERO-Long success. All variants use robot-video-pretrained
    initialization.
    }
    \label{tab:mot-ablation}
    \small
    \begin{tabular}{@{}lcc@{}}
        \toprule
        \textbf{Architecture}
        & \textbf{VTA (\%)}
        & \textbf{Joint (\%)} \\
        \midrule
        Non-MoT
        & 92.8
        & 93.6 \\
        MoT
        & \textbf{97.8}
        & \textbf{96.6} \\
        \bottomrule
    \end{tabular}
\end{table}

\begin{table*}[t]
    \centering
    \caption{
    Policy and dynamics accuracy on LIBERO-Long.
    LC-IDM and LC-FDM are independently trained specialist checkpoints;
    the multi-objective model trains policy, IDM, and FDM within a single
    checkpoint. 
    Lower is better for MSE and IDM error; higher is better for success and SSIM.
    }
    \label{tab:external-dynamics}
    \scriptsize
    \setlength{\tabcolsep}{3.2pt}
    \resizebox{\textwidth}{!}{
    \begin{tabular}{@{}l
        c
        cccc
        cccc@{}}
        \toprule
        &
        \textbf{Policy}
        &
        \multicolumn{4}{c}{\textbf{Real demonstrations}}
        &
        \multicolumn{4}{c}{\textbf{Counterfactual transitions}} \\
        \cmidrule(lr){2-2}
        \cmidrule(lr){3-6}
        \cmidrule(lr){7-10}

        \textbf{Model}
        & \textbf{Succ. (\%)}
        & \textbf{FDM MSE $\downarrow$}
        & \textbf{FDM SSIM $\uparrow$}
        & \textbf{IDM pos. (cm) $\downarrow$}
        & \textbf{IDM rot. ($^\circ$) $\downarrow$}
        & \textbf{FDM MSE $\downarrow$}
        & \textbf{FDM SSIM $\uparrow$}
        & \textbf{IDM pos. (cm) $\downarrow$}
        & \textbf{IDM rot. ($^\circ$) $\downarrow$} \\
        \midrule

        \ours, LC-IDM, CF-only
        & --
        & --
        & --
        & 1.76
        & 2.39
        & --
        & --
        & \textbf{1.17}
        & 3.60 \\

        \ours, LC-IDM, mixed
        & --
        & --
        & --
        & 0.44
        & 1.21
        & --
        & --
        & 1.22
        & 3.68 \\

        \ours, LC-FDM, CF-only
        & --
        & 0.00734
        & 0.9126
        & --
        & --
        & \textbf{0.00940}
        & \textbf{0.8889}
        & --
        & -- \\

        \ours, LC-FDM, mixed
        & --
        & 0.00186
        & 0.9776
        & --
        & --
        & 0.00962
        & 0.8859
        & --
        & -- \\

        \midrule

        \ours, multi-objective
        & \textbf{92.8}
        & \textbf{0.00058}
        & \textbf{0.9934}
        & \textbf{0.43}
        & \textbf{1.10}
        & 0.01416
        & 0.8398
        & 1.73
        & \textbf{3.55} \\

        UVA
        & 88.0
        & 0.00279
        & 0.9645
        & 2.03
        & 1.60
        & 0.01514
        & 0.8134
        & 3.57
        & 12.99 \\

        \bottomrule
    \end{tabular}
    }
\end{table*}

\subsection{Effect of the Mixture-of-Transformers Architecture}
\label{app:mot-ablation}

We isolate the effect of the Mixture-of-Transformers (MoT) parameterization
while keeping the pretrained visual foundation fixed. Both variants start
from the same robot-video-pretrained video checkpoint and use the same
downstream data and training procedure. In the non-MoT variant, video and
action tokens are processed by the same pretrained DiT; the MoT variant retains the pretrained video expert and introduces a separate action expert initialized from width-adapted copies of the corresponding video-expert layers. As shown in Table~\ref{tab:mot-ablation}, the
shared-DiT model already achieves strong performance, reaching 92.8\% for
VTA and 93.6\% for Joint. MoT further improves success to 97.8\% and
96.6\%, corresponding to gains of 5.0 and 3.0 percentage points,
respectively. 

\section{External Dynamics Comparison}
\label{app:external-dynamics}

We compare \ours with UVA~\cite{li2503unified}, the closest released
LIBERO system in our related work that exposes policy, inverse-dynamics,
and forward-dynamics objectives within one model. Table~\ref{tab:external-dynamics}
covers two settings: independently trained local-context dynamics components
and a single multi-objective checkpoint trained jointly across policy, IDM,
and FDM objectives.

For FDM, we report RGB prediction error in a common agent-view image space.
For IDM, the two systems use different action parameterizations, so we compare
the resulting end-effector trajectory error after executing the predicted
actions from the same simulator state.

\subsection{Independent Dynamics Components}

The LC-IDM and LC-FDM specialists are trained either on counterfactual
transitions only or on a 60/40 mixture of counterfactual and demonstration
transitions. CF-only specialists are strongest on the held-out
counterfactual distribution, while mixed supervision substantially improves
fidelity on real demonstrations. Both remain well ahead of UVA on the
counterfactual dynamics metrics.

\newpage

\subsection{Unified Policy and Dynamics Modeling}

UVA trains policy generation, inverse dynamics, and forward dynamics within a
single model. We therefore also evaluate one \ours checkpoint trained with
60\% real joint video--action supervision, 20\% local-context IDM supervision,
and 20\% local-context FDM supervision; the IDM and FDM portions are each split
equally between demonstration and counterfactual transitions.

The unified \ours checkpoint outperforms UVA on policy success and on the
evaluated IDM and FDM metrics. Its counterfactual dynamics accuracy is weaker
than that of the dedicated specialists, showing the expected tradeoff between
multi-objective training and specialization. The unified \ours checkpoint also gives the lowest IDM and FDM errors on real
demonstrations. These numbers measure fit to the demonstration distribution,
not held-out generalization, because some of the same demonstrations may also
be seen by the policy objective during training. Its policy score is also lower
than the policy-specialized \ours models as only 60\% of training updates
optimize the policy objective.

\end{document}